\documentclass[journal,twoside,web]{ieeecolor}
\usepackage{generic}
\usepackage{cite}
\usepackage{amsmath,amssymb,amsfonts}
\usepackage{algorithmic}
\usepackage{graphicx}
\usepackage{algorithm}
\usepackage{hyperref}
\hypersetup{hidelinks=true}
\usepackage{textcomp}
\usepackage{booktabs}
\usepackage{multirow}
\usepackage{array}
\usepackage{tabularx}

\def\BibTeX{{\rm B\kern-.05em{\sc i\kern-.025em b}\kern-.08em
    T\kern-.1667em\lower.7ex\hbox{E}\kern-.125emX}}
\begin{document}
\title{Sequence Search: Automated Sequence Design using Neural Architecture Search}
\author{Rokgi Hong, Hongjun An, Sooyeon Ji, and Jongho Lee, \IEEEmembership{Member, IEEE}
\thanks{This research was supported by Korean Government Grants: RS-2024-00439677, RS-2024-00438392, IITP-2026-RS-2023-00256081, and RS-2024-00435727. This work was also supported by the Institute of New Media and Communications, the Institute of Engineering Research, and the Electric Power Research Institute at Seoul National University, as well as the Hankuk University of Foreign Studies Research Fund of 2025.}
\thanks{Corresponding authors: Sooyeon Ji and Jongho Lee.}
\thanks{Rokgi Hong, Hongjun An, and Jongho Lee are with the Department of Electrical and Computer Engineering, Seoul National University, Seoul, South Korea (e-mail: hrocky125@snu.ac.kr; plynt@snu.ac.kr; jonghoyi@snu.ac.kr).}
\thanks{Sooyeon Ji is with the Division of Computer Engineering, Hankuk University of Foreign Studies, Yongin, South Korea (e-mail: sueji0221@hufs.ac.kr).}}

\maketitle

\begin{abstract}
Developing an MR sequence is challenging and remains largely constrained by human intuition. Recently, AI-driven approaches have been proposed; however, most require an initial sequence for parameter optimization or extensive training datasets, limiting their general applicability. In this study, we propose ``Sequence Search,'' an automated sequence design framework based on neural architecture search. The method takes tissue properties, imaging parameters, and design objectives as inputs and generates pulse sequences satisfying the design objectives, without requiring prior knowledge of conventional sequence structures. Sequence Search iteratively generates candidate sequences through neural architecture search and optimizes them via a differentiable Bloch simulator and objective-specific loss functions using gradient-based learning. The framework successfully replicated conventional spin-echo, T\textsubscript{2}-weighted spin-echo, and inversion recovery sequences. Less intuitive solutions were also discovered, such as three-RF spin-echo-like sequences with reduced RF energy and refocusing phases deviating from the conventional Hahn-echo. This work establishes a generalizable framework for automated MR sequence design, highlighting the potential to explore configurations beyond conventional designs based on human intuition.
\end{abstract}

\section{Introduction}
\IEEEPARstart{M}{agnetic} resonance imaging (MRI) is known for its flexibility in creating various image contrasts, obtained through MRI pulse sequences---schedules of radio-frequency (RF) pulses and linear gradient magnetic fields along the three axes.

So far, sequence design has relied on experts' knowledge and intuition, requiring a deep understanding of MR physics and spin dynamics. The non-linearity of the Bloch equation and system imperfections such as $B_0$ and $B_1$ field inhomogeneity further complicate this process, constraining current sequences to human intuition.

Recently, AI-powered designs have demonstrated novel results beyond human intuition in areas such as protein engineering \cite{bib1} and drug discovery \cite{bib2}. In the field of MRI sequence design, AI has first been applied to design individual sequence components---including RF and/or gradient waveforms\cite{bib3, bib4, bib5, bib6, bib7, bib8} and k-space acquisition order\cite{bib9, bib10, bib11, bib12, bib13}---delivering unconventional designs (e.g., inversion pulse design from DeepRF \cite{bib3}) and demonstrating potential. Beyond component-level design, other studies have optimized sequence-level parameters---the number, timing (echo time; TE, repetition time; TR), and properties (e.g., flip angle; FA) of RF and gradient pulses---using optimization algorithms\cite{bib14} or supervised learning jointly with reconstruction networks\cite{bib15, bib16}. A deep reinforcement learning approach has also been developed to find the phase encoding step, TE, TR, and FA of a sequence for deducing the geometric shape of an object in the scanner\cite{bib17}. However, these frameworks rely on training datasets or start from an initial sequence consisting of a specific number of sequence components (e.g., RFs, gradient, read-out) in a specific arrangement, and optimize their parameters, limiting their flexibility in discovering a new sequence.

\begin{figure*}[!t]
\centerline{\includegraphics[width=0.8\textwidth]{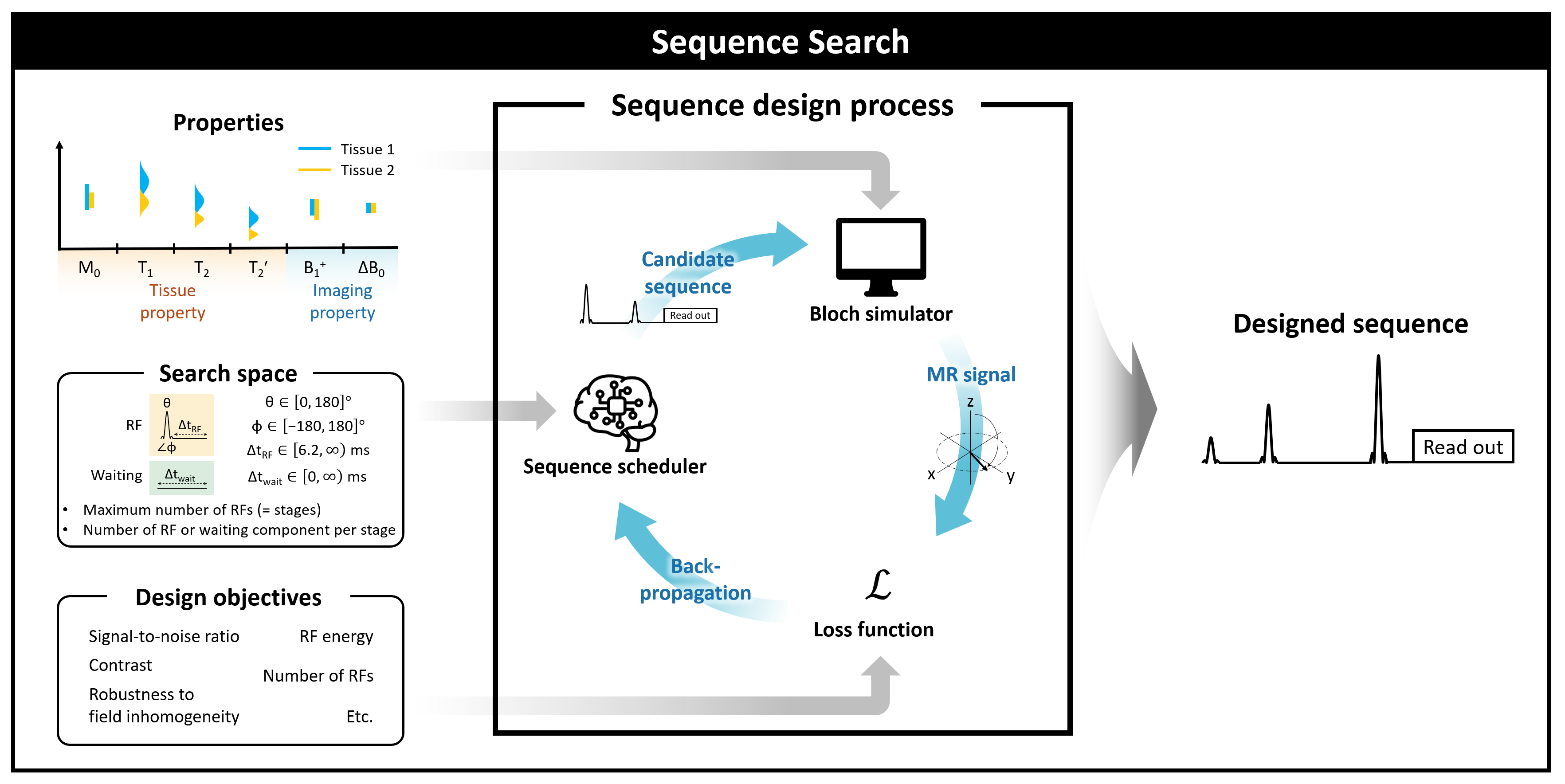}}
\caption{Overview of Sequence Search. Sequence Search designs a sequence within the given search space based on the input properties and design objectives, without requiring prior knowledge of sequence design principles. Sequence Search consists of three main parts. The sequence scheduler recommends a candidate sequence. The Bloch simulator processes this sequence in conjunction with the given properties, generating MR signals to calculate the loss function derived from objectives. The loss function is back-propagated for optimization of the sequence scheduler.
}
\label{fig1}
\end{figure*}

In machine learning, automated machine learning has emerged as a powerful tool for designing an AI model\cite{bib18}. Neural architecture search (NAS)\cite{bib19}, a subfield of automated machine learning, automates the design of neural network structures and has produced architectures surpassing hand-crafted designs\cite{bib20}.

In this study, we introduce a novel sequence design methodology that automatically generates an MR pulse sequence for a given objective. We draw a principled analogy between NAS and MR pulse sequence design, and leverage it to develop ``Sequence Search''---a framework that searches over the space of pulse sequence configurations to identify optimal sequences for given design objectives. Sequence Search designs a sequence based on design objectives and tissue and imaging properties without any prior knowledge of MR sequence design or training datasets, differentiating it from the previously proposed methods. Furthermore, Sequence Search simultaneously optimizes sequence parameters and structure, providing distinct flexibility in the designed sequence.

\newpage
\section{Methods}
\subsection{Overview of Sequence Search}

Sequence Search aims to generate an MR sequence that satisfies predefined design objectives under specified tissue (e.g., $M_0$, $T_1$, $T_2$ and $T_2^\prime$) and imaging properties (e.g., $B_1^+$ and $\Delta B_0$). The framework consists of three components (Fig. \ref{fig1}): a sequence scheduler, a Bloch simulator, and a loss function (see Section II.B--II.D for details). The optimization proceeds iteratively. First, the sequence scheduler proposes a candidate sequence. Next, the Bloch simulator computes the resulting MR signals based on the proposed sequence and the specified tissue and imaging properties. The loss function, defined according to the design objectives, is then evaluated from the simulated signals. The computed loss is back-propagated to update the sequence scheduler, enabling generation of an improved candidate sequence in the subsequent iteration.

In this study, the design space is restricted to RF pulse scheduling with at most five RF pulses. Specifically, the optimization includes the timing, flip angle, and phase of each RF pulse, which are critical determinants of many MRI contrast mechanisms. To reduce the search space dimension, the gradient waveforms and readout scheme were not optimized in this work. TR was assumed to be infinite (i.e., full recovery), disregarding steady-state effects. RF pulses were modeled as instantaneous rotations, and signal readout was assumed to occur instantaneously.

\subsection{Sequence scheduler}
The sequence scheduler determines the number of RF pulses, the flip angle and phase of each RF pulse, and the timing between RF pulses and/or the readout. The sequence scheduler is based on NAS, originally developed for automated neural network design by identifying an optimal architecture within a predefined search space \cite{bib20}. In a conventional NAS, the search space consists of candidate operations (e.g., convolution, pooling) and possible network topologies, such as depth and connectivity. NAS therefore determines not only which operations to select but also how to connect them, often using directed acyclic graphs or modular cell structures.

\begin{table*}[!t]
\centering
\footnotesize
\caption{Tissue properties and imaging properties for the experiments}
\label{table1}
\begin{tabular}{c c c c c c c c}
\toprule
& \multicolumn{4}{c}{Tissue properties} & \multicolumn{3}{c}{Imaging properties} \\
\cmidrule(lr){2-5} \cmidrule(lr){6-8}
Tissue 
& $M_0$ 
& $T_1$ (ms) 
& $T_2$ (ms) 
& $T_2^{\prime}$ (ms) 
& $B_1^+$ 
& $\Delta B_0$ (Hz) 
& Intra-voxel frequency spread (Hz) \\
\midrule

GM 
& 0.8 {\textpm} 0.02
& 1,331 {\textpm} 57 
& 110 {\textpm} 9
& 170 {\textpm} 27 
& \multirow{3}{*}{0.8--1.2}
& \multirow{3}{*}{-50--50}
& \multirow{3}{*}{-30--30} \\

WM 
& 0.7 {\textpm} 0.02
& 832 {\textpm} 44
& 80 {\textpm} 3
& 161 {\textpm} 17 
& & & \\

CSF 
& 1.0
& 4,000 {\textpm} 200 
& 2,000 {\textpm} 100
& 6,000 {\textpm} 300
& & & \\

\bottomrule
\end{tabular}
\end{table*}

\begin{figure}[!b]
\centerline{\includegraphics[width=\columnwidth]{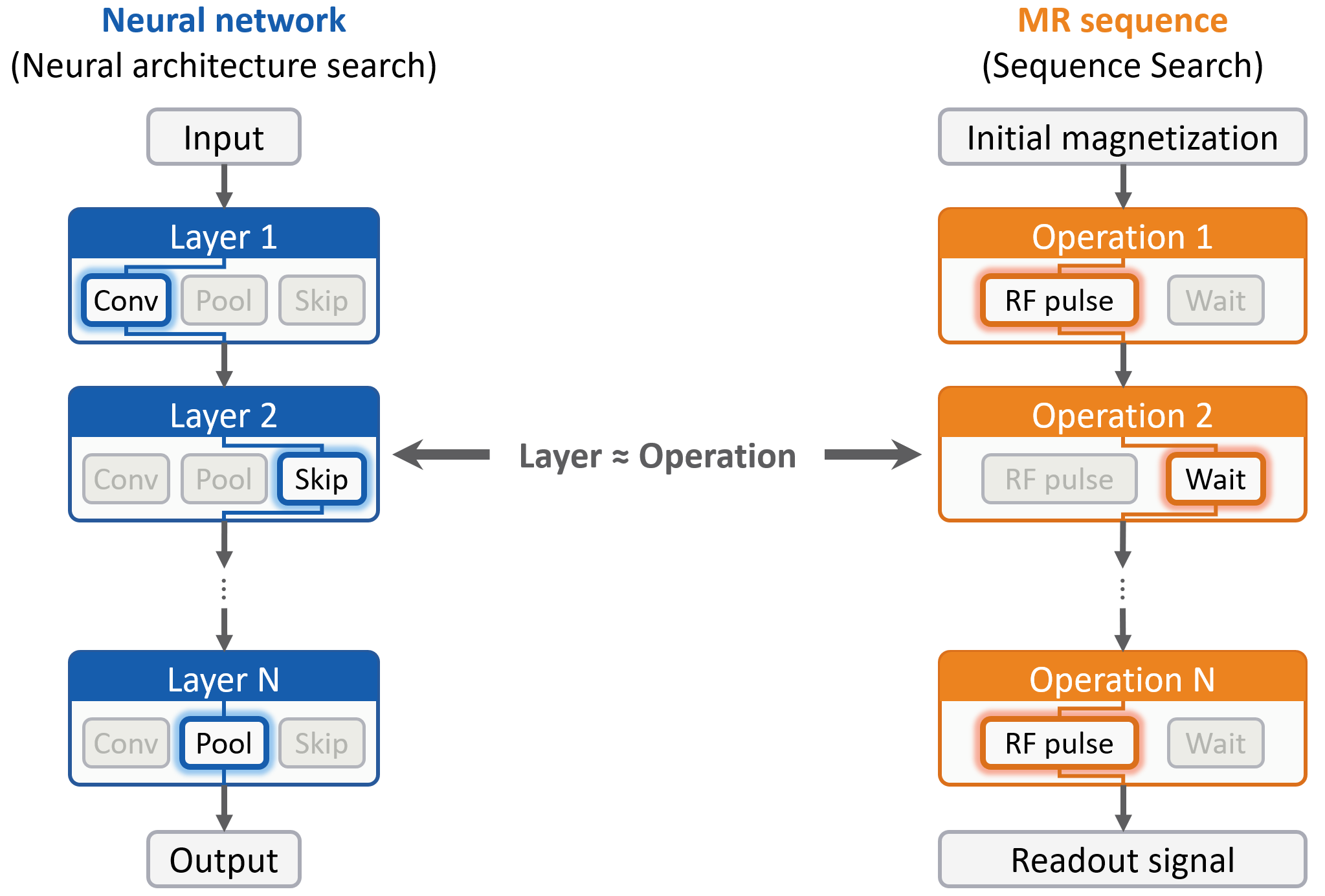}}
\caption{Structural analogy between NAS and MR sequence design. In NAS, a network is built by selecting operations for each layer from a predefined set. Similarly, in the proposed framework, an MR sequence is represented as a stack of operations, where each layer corresponds to a selectable sequence operation (RF pulse or waiting period) parameterized by flip angle, phase, or idle time.}
\label{fig2}
\end{figure}

Among various NAS strategies, we adopted ProxylessNAS \cite{bib21}, a gradient-based method that is computationally efficient compared to other search strategies such as reinforcement learning or evolutionary search. Unlike conventional NAS methods that explore diverse network topologies, ProxylessNAS operates on a fixed number of layers, where each layer is associated with a small set of candidate operations. Architecture parameters determine which operation is selected at each layer, enabling joint optimization of architecture parameters and weight parameters through gradient descent \cite{bib21}. This formulation aligns with MR sequence design: each RF pulse or waiting period corresponds to a selectable operation, parameterized by flip angle, phase, and idle time, and a complete sequence is represented as a stack of layers (Fig. \ref{fig2}).

\begin{figure}[!b]
\centerline{\includegraphics[width=\linewidth]{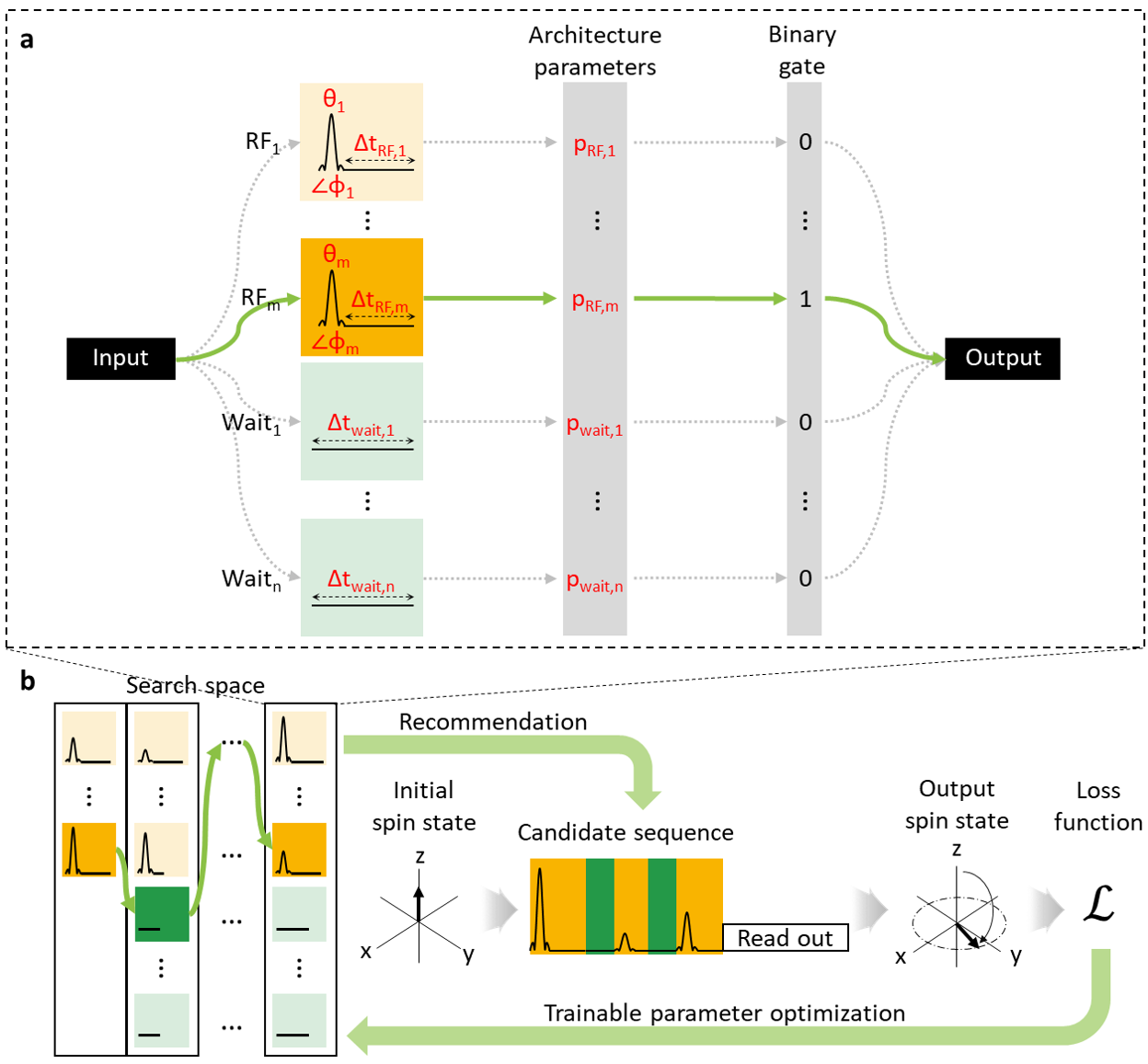}}
\caption{Overview of the sequence scheduler using NAS. (a) Operations and trainable parameters (highlighted in red) for a single layer of search space. In this work, each layer selects one operation out of five candidate RF operations and three candidate wait operations (corresponding to m=5, n=3 in the figure). (b) Using NAS, a candidate sequence is generated from the search space. The resulting spin state is used to compute the loss function for optimization.
}
\label{fig3}
\end{figure}

The search space consisted of five layers, each selecting one operation from five candidate RF operations and three candidate waiting operations, allowing for a maximum of five RF pulses per sequence (Fig. \ref{fig3}). The first layer was constrained to begin with an RF pulse. Initialization was structured to encourage exploration across the parameter space. RF flip angles were initialized within compartmentalized ranges spanning 0--180{\textdegree} in 36{\textdegree} intervals, and RF phases were randomly initialized between -180{\textdegree} and 180{\textdegree}. Idle times were initialized to 6.2 ms for RF operations and within three logarithmically spaced ranges (1--10, 10--100, 100--1000 ms) for waiting operations. The idle time after an RF pulse ($\Delta t_\textrm{RF}$) was constrained to be at least 6.2 ms to prevent unrealistically rapid RF pulse succession. Architecture parameters that govern operation selection at each layer were initialized from a Gaussian distribution with mean 0 and standard deviation 0.001.

To stabilize optimization, idle times ($\Delta t_{\textrm{RF}}$ and $\Delta t_{\textrm{wait}}$ in Fig. \ref{fig3}) were log-transformed to compress their wide dynamic range. A small constant of 0.001 ms was added prior to logarithmic transformation to prevent numerical instability when idle times approached zero.

Weight parameters of the sequence operations were optimized using SGD (learning rate 0.01), and architecture parameters using ADAM \cite{bib22} (learning rate 0.001).

\subsection{Bloch simulator}
A Bloch simulator was implemented to model the MR signal evolution within a voxel while accounting for $T_1$, $T_2$, and $T_2^\prime$ relaxation, as well as $B_0$ and $B_1^+$ field inhomogeneities. $T_1$ and $T_2$ relaxation effects were applied during idle periods of the sequence. $B_1^+$ inhomogeneity was incorporated by scaling the flip angle of each RF pulse by a voxel-specific $B_1^+$ factor. Voxel-wise $B_0$ inhomogeneity ($\Delta B_0$) was modeled as a constant frequency offset causing cumulative phase accrual over time. To capture intra-voxel dephasing, each voxel was represented by 256 spins with additional frequency offsets from $T_2^\prime$-related microscopic frequency dispersion (Lorentzian) \cite{bib23} and an intra-voxel frequency spread (linear) \cite{bib24}.

To approximate whole-brain variability, 100,000 synthetic voxels were generated by sampling tissue and imaging properties (Table \ref{table1}). Three major brain tissue types---gray matter (GM), white matter (WM), and cerebrospinal fluid (CSF)---were modeled, with approximately equal numbers of voxels assigned to each. Tissue parameters for GM, WM, and CSF were adapted from prior literature \cite{bib25, bib26, bib27, bib28}. $B_1^+$ values were sampled uniformly between 0.8 and 1.2, and $B_0$ ranged from -50 to 50 Hz. Intra-voxel frequency spread was modeled by assigning frequency offsets uniformly distributed between -30 and 30 Hz across the constituent spins within each voxel. Optimization was conducted with a batch size of 1,000 voxels per iteration.

To enable gradient-based optimization, the Bloch simulator was implemented as a differentiable module using automatic differentiation in the PyTorch library \cite{bib29}. The full computational graph was constructed to permit back-propagation through the entire sequence simulation process.

\subsection{Loss functions}
The following loss functions were defined and combined according to the design objective to optimize the sequence scheduler.

A signal intensity loss, $\mathcal{L}_\textrm{sig}$, was defined to maximize signal intensity as follows:
\begin{equation} \mathcal{L}_\textrm{sig} = \left\Vert M_0 - M_{xy}\right\Vert^2_2,\label{eq1}\end{equation}
where $M_0$ denotes fully relaxed magnetization, $M_{xy}$ denotes the transverse signal at read-out.

A signal nulling loss, $\mathcal{L}_\textrm{null}$, was defined to suppress signal intensity:
\begin{equation} \mathcal{L}_\textrm{null} = \left\Vert M_{xy}\right\Vert^2_2.\label{eq2}\end{equation}

To enhance contrast between tissue $A$ and tissue $B$, a contrast loss, $\mathcal{L}_{\textrm{cont;}A,B}$, was defined as:
\begin{equation} \mathcal{L}_{\textrm{cont;}A,B} = -\left\Vert M_{xy, A} - M_{xy, B}\right\Vert^2_2,\label{eq3}\end{equation}
where $M_{xy, A}$ and $M_{xy, B}$ denote the transverse signal at read-out for tissue A and tissue B, respectively.

Additionally, we defined RF number loss ($\mathcal{L}_\textrm{RFnumber}$) to prevent excessive usage of RF pulses and RF energy loss ($\mathcal{L}_\textrm{RFenergy}$) to reduce RF power deposition. RF number loss was defined as the total number of RF pulses in the sequence, and RF energy loss was defined as the sum of the squares of the flip angles.

\subsection{Experiments}
To evaluate the performance of Sequence Search, we conducted three experiments targeting three distinct objectives: (i) maximizing signal intensity in the presence of intra-voxel $B_0$ field inhomogeneity, (ii) enhancing GM--WM contrast, and (iii) CSF nulling. Each experiment was formulated with a corresponding loss function, as described in the following subsections.

To assess reproducibility, each experiment was repeated 100 times using different random seeds. Designs whose loss values were within 5\% of the optimal sequence identified by grid search (Section II.E.4) were defined as successful designs and subjected to categorization. The sequences were categorized according to their structural characteristics. All remaining designs were grouped under ``Others.'' Sequence categorization was based solely on the structural configuration of RF pulses---particularly their flip angles and phase relationships---rather than on performance metrics such as signal intensity, contrast, or RF energy. Successful designs that did not satisfy any predefined structural category were also assigned to Others.

RF pulses with flip angles smaller than 3{\textdegree} were removed from the sequence prior to categorization and evaluation, as their contribution to the resulting signal was considered negligible. An RF phase standardization process was applied prior to categorization: (i) all phases were shifted such that the first RF phase equals 0{\textdegree} to remove global phase offsets, (ii) phases were normalized to the range [-180{\textdegree}, 180{\textdegree}], and (iii) all phases were sign-inverted when the second RF phase was negative to enforce consistent phase orientation across sequences.

After categorization, the mean and standard deviation of the sequence parameters (RF flip angle, RF phase, time from first RF pulse to readout) and the resulting signal characteristics (GM, WM, and CSF signal intensity) were calculated. When computing the RF phase statistics, the standardization was adapted to the phase distribution to avoid bias arising from phase wrap-around (e.g., omitting the sign-inversion step when phases clustered near 0{\textdegree} or {\textpm}180{\textdegree}, and shifting negative values by 360{\textdegree} for continuity around 180{\textdegree}). 

\subsubsection{Experiment 1: Maximize signal intensity}
This experiment aimed to design a pulse sequence that maximizes signal intensity in the presence of intra-voxel $B_0$ field inhomogeneity, which induces spin dephasing. The resulting sequences were expected to reproduce the signal refocusing behavior of a spin-echo sequence.

To reflect realistic imaging conditions, voxel-wise off-resonance ($\Delta B_0$) and $B_1^+$ transmit field inhomogeneity were also incorporated into the simulation. While $\Delta B_0$ causes time-dependent phase accrual due to off-resonance effects, $B_1^+$ variations affect flip angle accuracy; both factors can degrade refocusing performance. Robustness to these factors was enforced by explicitly simulating a broad range of $\Delta B_0$ and $B_1^+$ values across the training voxels. In addition to signal maximization, penalties on RF energy and the number of RF pulses were included to discourage energetically inefficient or structurally excessive solutions. These objectives were encoded in the following loss function:
\begin{equation} \mathcal{L} = \lambda_\textrm{sig}\mathcal{L}_\textrm{sig} + \lambda_\textrm{RFenergy}\mathcal{L}_\textrm{RFenergy} + \lambda_\textrm{RFnumber}\mathcal{L}_\textrm{RFnumber}, \label{eq4}\end{equation}
where  $\lambda_\textrm{sig}$, $\lambda_\textrm{RFenergy}$, and $\lambda_\textrm{RFnumber}$ were set to 1, 0.0001 and 0.001, respectively.

The resulting sequences were categorized according to the number of RF pulses and the properties of the final two RF pulses, as summarized in Table \ref{table2}. Specifically, sequences were classified as Hahn-echo-like sequences when the flip angles of the last two RF pulses were approximately 90{\textdegree}--180{\textdegree} and the refocusing phase was near 0{\textdegree} or 180{\textdegree}. Sequences with similar flip angles but different phase relationships were classified as other spin-echo-like sequences. For sequences containing three or more RF pulses, earlier pulses were not considered in determining the sequence category if their flip angles were smaller than 20{\textdegree}, as such pulses were interpreted as preparatory or negligible perturbations rather than defining refocusing elements. All remaining designs were grouped as Others.

\begin{table}[!t]
\centering
\footnotesize
\caption{Classification criteria for signal intensity maximization experiment}
\label{table2}
\setlength{\tabcolsep}{0pt}

\begin{tabular}{
    >{\centering\arraybackslash}m{0.34\linewidth}
    >{\centering\arraybackslash}m{0.22\linewidth}
    >{\centering\arraybackslash}m{0.16\linewidth}
    >{\centering\arraybackslash}m{0.28\linewidth}
}
\toprule
Category
& Flip angles of last two RFs
& Refocusing RF phase
& Flip angles of early RFs (if any) \\
\midrule

Hahn-echo-like sequences
& 90{\textdegree}--180{\textdegree}
& 0{\textdegree} or {\textpm}180{\textdegree}
& {\textless} 20{\textdegree} \\

Other spin-echo-like sequences
& 90{\textdegree}--180{\textdegree}
& Otherwise
& {\textless} 20{\textdegree} \\

\midrule

Tolerance
& {\textpm}20{\textdegree}
& {\textpm}20{\textdegree}
& -- \\
\bottomrule
\end{tabular}
\end{table}

\subsubsection{Experiment 2: Enhance GM--WM contrast}
This experiment aimed to design a pulse sequence that enhances GM--WM signal contrast. To explicitly encourage contrast between the two tissues, an additional loss term was introduced to penalize small signal differences between GM and WM. The final loss function was defined as:
\begin{equation} \begin{aligned}
\mathcal{L} = &\lambda_\textrm{sig}\mathcal{L}_\textrm{sig} + \lambda_\textrm{cont}\mathcal{L}_\textrm{cont;GM,WM} \\&+ 
\lambda_\textrm{RFenergy}\mathcal{L}_\textrm{RFenergy} + \lambda_\textrm{RFnumber}\mathcal{L}_\textrm{RFnumber}, \label{eq5}
\end{aligned}\end{equation}
where $\lambda_\textrm{sig} = 1$, $\lambda_\textrm{cont} = 30$, $\lambda_\textrm{RFenergy} = 0.0001$, and $\lambda_\textrm{RFnumber} = 0.001$.

To isolate the effect of $T_2$ weighting from $T_1$-driven contrast, the experiment was first conducted under a controlled condition in which GM and WM were assigned identical $T_1$ values (1,000 {\textpm} 50 ms), while all other tissue and imaging parameters were kept unchanged. The experiment was then repeated using the tissue-specific $T_1$ values listed in Table \ref{table1} to evaluate how $T_1$ differences influence the optimization outcome. To further assess the sensitivity of the optimization to the contrast objective, additional experiments were performed under the identical $T_1$ value condition with modified contrast weights of $\lambda_\textrm{cont} = 20$ and 50, while keeping the remaining weights fixed.

The resulting sequences were categorized using the same structural criteria as in Experiment 1, based on the flip angles and phase difference of the last two RF pulses. In experiments with distinct GM and WM $T_1$ values, an additional category, inversion-prepared spin-echo-like sequences, was introduced. These sequences were defined by the presence of an initial inversion pulse with a flip angle within the range 180{\textdegree} {\textpm} 20{\textdegree}, followed by a spin-echo readout (e.g., 180{\textdegree}--90{\textdegree}--180{\textdegree} configuration), and an inversion time (TI) greater than 100 ms.

\subsubsection{Experiment 3: CSF nulling}
The objective of this experiment was to design a sequence that selectively nulls the CSF signal while preserving signal from GM and WM. Therefore, the designed sequences were expected to emulate the functional behavior of an inversion recovery sequence by selectively nulling the CSF signal. To achieve this objective, the loss function was defined as:
\begin{equation}\begin{aligned} \mathcal{L} = & \lambda_\textrm{sig}\left(\mathcal{L}_\textrm{sig;GM} + \mathcal{L}_\textrm{sig;WM}\right) + \lambda_\textrm{null}\mathcal{L}_\textrm{null;CSF} \\&+ \lambda_\textrm{RFenergy}\mathcal{L}_\textrm{RFenergy} + \lambda_\textrm{RFnumber}\mathcal{L}_\textrm{RFnumber}. \label{eq6}\end{aligned}\end{equation}

Here, $\mathcal{L}_\textrm{sig;GM}$ and $\mathcal{L}_\textrm{sig;WM}$ penalize signal loss in GM and WM, while $\mathcal{L}_\textrm{null;CSF}$ penalizes residual signal in CSF. The coefficients $\lambda_\textrm{sig}$, $\lambda_\textrm{null}$, $\lambda_\textrm{RFenergy}$, and $\lambda_\textrm{RFnumber}$ were set to 0.75, 0.25, 0.0001, and 0.001, respectively.

To evaluate the sensitivity of the optimization to the CSF suppression objective, the experiment was repeated with an increased CSF nulling weight ($\lambda_\textrm{null} = 1.0$), while keeping the remaining coefficients fixed.

The resulting sequences were categorized according to the flip angles and phase relationships of the relevant RF pulses, together with TI greater than 2.7 s. Two principal categories were defined (Table \ref{table3}): (i) inversion-prepared gradient-echo-like sequences (180{\textdegree}--90{\textdegree} configuration) and (ii) inversion-prepared spin-echo-like sequences (180{\textdegree}--90{\textdegree}--180{\textdegree} configuration). Sequences that did not meet these criteria were classified as Others.

\begin{table}[!t]
\centering
\footnotesize
\caption{Classification criteria for CSF nulling experiment}
\label{table3}
\setlength{\tabcolsep}{0pt}
\begin{tabular}{
    >{\centering\arraybackslash}m{0.65\linewidth}
    >{\centering\arraybackslash}m{0.25\linewidth}
    >{\centering\arraybackslash}m{0.10\linewidth}
}
\toprule
Category & Flip angles & TI \\
\midrule

Inversion-prepared gradient-echo-like sequences
& 180{\textdegree}--90{\textdegree}
& {\textgreater} 2.7 s \\

Inversion-prepared spin-echo-like sequences
& 180{\textdegree}--90{\textdegree}--180{\textdegree}
& {\textgreater} 2.7 s \\

\midrule

Tolerance
& {\textpm}20{\textdegree}
& -- \\
\bottomrule
\end{tabular}
\end{table}

\subsubsection{Grid search}
For each experiment, a grid search over a two-RF-pulse sequence was performed as a reference benchmark. The search was restricted to parameter neighborhoods around conventional configurations (90{\textdegree}--180{\textdegree} spin-echo or 180{\textdegree}--90{\textdegree} inversion recovery), optimizing $\lbrace\theta_1, \theta_2, \phi_2, \Delta t_1, \Delta t_2\rbrace$ with $\phi_1$ fixed at 0{\textdegree}. The parameter ranges for each experiment are summarized in Table \ref{table4}. Tissue parameters, field inhomogeneity conditions, and loss functions were identical to those used in the Sequence Search experiments. The total number of grid combinations ranged from 2.7 {\texttimes} 10\textsuperscript{6} to 9.7 {\texttimes} 10\textsuperscript{6} across experiments.

\begin{table}[!t]
\centering
\footnotesize
\caption{Parameter ranges for grid search}
\label{table4}
\renewcommand{\arraystretch}{1.00}
\setlength{\tabcolsep}{0pt}

\begin{tabular}{@{}
>{\centering\arraybackslash}m{0.17\columnwidth}
>{\centering\arraybackslash}m{0.29\columnwidth}
>{\centering\arraybackslash}m{0.27\columnwidth}
>{\centering\arraybackslash}m{0.27\columnwidth}
@{}}
\toprule
Parameter &
Experiment 1
(Signal maximization) &
Experiment 2
(GM-WM contrast) &
Experiment 3
(CSF nulling) \\
\midrule

Configuration &
90{\textdegree}--180{\textdegree} &
90{\textdegree}--180{\textdegree} &
180{\textdegree}--90{\textdegree} \\

$\theta_1$ &
85{\textdegree}--95{\textdegree} (1{\textdegree}) &
85{\textdegree}--95{\textdegree} (1{\textdegree}) &
170{\textdegree}--180{\textdegree} (1{\textdegree}) \\

$\theta_2$ &
170{\textdegree}--180{\textdegree} (1{\textdegree}) &
170{\textdegree}--180{\textdegree} (1{\textdegree}) &
85{\textdegree}--95{\textdegree} (1{\textdegree}) \\

$\phi_2$ &
0{\textdegree}--180{\textdegree} (1{\textdegree}) &
0{\textdegree}--180{\textdegree} (1{\textdegree}) &
0{\textdegree}--180{\textdegree} (5{\textdegree}) \\

$\Delta t_1$ &
6.2--7.2 ms (0.1) &
15--25 ms (0.5) &
2.78--3.80 s (0.01) \\

$\Delta t_2$ &
6.2--7.2 ms (0.1) &
15--25 ms (0.5) &
6.2--11.2 ms (1.0) \\

\midrule
Total combinations &
$2.7 \times 10^6$ &
$9.7 \times 10^6$ &
$2.8 \times 10^6$ \\
\bottomrule
\end{tabular}
\end{table}

\subsection{Optimization details}
The entire procedure was implemented in Python using the PyTorch \cite{bib29} library. All experiments were conducted on a GPU workstation equipped with an NVIDIA GeForce RTX 5090 GPU and an AMD Ryzen Threadripper 7960X CPU (4.2 GHz). For each sequence search experiment, optimization was performed for 1,000 epochs, with each run requiring approximately 2.2 hours of computation. The source code is publicly available at https://github.com/SNU-LIST/SequenceSearch.

\section{Results}
\subsection{Experiment 1: Maximize signal intensity}
\begin{figure}[!b]
\centerline{\includegraphics[width=\linewidth]{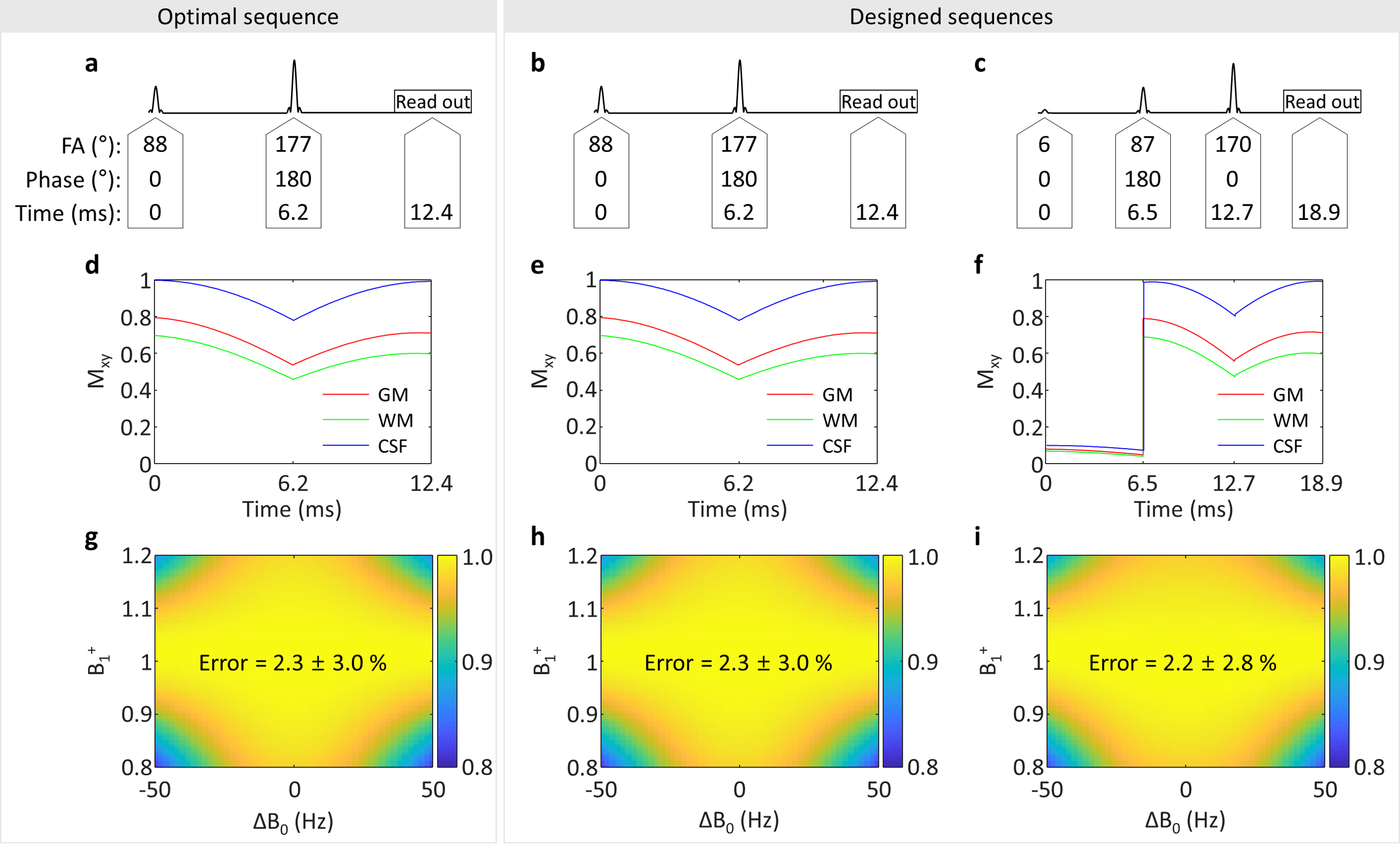}}
\caption{Results of the signal intensity maximization experiment. (a) Optimal two-RF sequence identified by grid search (88{\textdegree}--177{\textdegree}). (b, c) Designed sequences: (b) two-RF Hahn-echo-like sequence (88{\textdegree}--177{\textdegree}); (c) three-RF Hahn-echo-like sequence (6{\textdegree}--87{\textdegree}--170{\textdegree}). (d--f) Simulated transverse-magnetization evolution for GM, WM, and CSF corresponding to (a--c) at $B_1^+$=1 and $\Delta B_0$=0. (g--i) Robustness to field inhomogeneity: GM signal intensity over $\Delta B_0$ and $B_1^+$, normalized to the on-resonance value.
}
\label{fig4}
\end{figure}

In Experiment 1, Sequence Search generated two dominant spin-echo-like sequences (Table \ref{table5}): two-RF Hahn-echo-like sequences ([88.4 {\textpm} 0.1{\textdegree}]--[176.7 {\textpm} 0.2{\textdegree}]; 55\% occurrence rate; Fig. \ref{fig4}b) and three-RF Hahn-echo-like sequences incorporating a small preparatory flip angle (e.g., [6.1 {\textpm} 0.3{\textdegree}]--[87.4 {\textpm} 0.2{\textdegree}]--[170.1 {\textpm} 0.3{\textdegree}]; 11\% occurrence rate; Fig. \ref{fig4}c). Both designed sequence types successfully refocused the dephased magnetization and exhibited spin-echo behavior (Fig. \ref{fig4}e--f).

\begin{table*}[!t]
\centering
\footnotesize
\caption{Results of the signal-intensity maximization experiment}
\label{table5}
\setlength{\tabcolsep}{0pt}

\begin{tabular}{@{}
>{\centering\arraybackslash}m{0.10\textwidth}   
>{\centering\arraybackslash}m{0.21\textwidth}   
>{\centering\arraybackslash}m{0.09\textwidth}   
>{\centering\arraybackslash}m{0.05\textwidth}   
>{\centering\arraybackslash}m{0.05\textwidth}   
>{\centering\arraybackslash}m{0.05\textwidth}   
>{\centering\arraybackslash}m{0.10\textwidth}   
>{\centering\arraybackslash}m{0.16\textwidth}   
>{\centering\arraybackslash}m{0.09\textwidth}   
>{\centering\arraybackslash}m{0.10\textwidth}   
@{}}
\toprule

\multicolumn{2}{c}{\multirow{2}{*}{Category}} &
\multirow{2}{*}{\parbox[c]{0.09\textwidth}{\centering Occurrence rate (\%)}} &
\multicolumn{3}{c}{Signal intensity} &
\multirow{2}{*}{\parbox[c]{0.10\textwidth}{\centering Relative RF energy (\%)}} &
\multirow{2}{*}{\parbox[c]{0.16\textwidth}{\centering Time from the first RF to the readout (ms)}} &
\multirow{2}{*}{\parbox[c]{0.09\textwidth}{\centering Flip angles ({\textdegree})}} &
\multirow{2}{*}{\parbox[c]{0.10\textwidth}{\centering Refocusing RF phase ({\textdegree})}} \\

\cmidrule(lr){4-6}
& & & GM & WM & CSF & & & & \\

\midrule
\multicolumn{2}{c}{Conventional Hahn-echo sequence (90{\textdegree}--180{\textdegree})}
& -- & 0.72 & 0.60 & 0.99
& 100.0 & 12.4 & [90.0, 180.0] & 180.0 \\

\midrule

\multicolumn{2}{c}{Optimal grid-search sequence (88{\textdegree}--177{\textdegree})}
& -- & 0.71 & 0.60 & 0.99
& 96.3 & 12.4 & [88.0, 177.0] & 180.0 \\

\midrule

\multirow{3}{*}{\parbox[c]{0.10\textwidth}{\centering Successful designs}}
& Two-RF Hahn-echo-like sequences (88{\textdegree}--177{\textdegree})
& 55
& 0.71 {\textpm} 0.00 & 0.60 {\textpm} 0.00 & 0.99 {\textpm} 0.00
& 96.4 {\textpm} 0.2 & 12.4 {\textpm} 0.0
& [88.4 {\textpm} 0.1, 176.7 {\textpm} 0.2]
& 179.9 {\textpm} 0.3 \\

& Three-RF Hahn-echo-like sequences (6{\textdegree}--87{\textdegree}--170{\textdegree})
& 11
& 0.71 {\textpm} 0.00 & 0.60 {\textpm} 0.00 & 0.99 {\textpm} 0.00
& 90.4 {\textpm} 0.2 & 18.9 {\textpm} 0.3
& [6.1 {\textpm} 0.3, 87.4 {\textpm} 0.2, 170.1 {\textpm} 0.3]
& 180.0 {\textpm} 0.2 \\

\midrule

\multicolumn{2}{c}{Others}
& 34
& 0.60 {\textpm} 0.07 & 0.50 {\textpm} 0.05 & 0.84 {\textpm} 0.10
& 82.0 {\textpm} 67.7 & 17.8 {\textpm} 9.2
& -- & -- \\

\bottomrule
\end{tabular}
\end{table*}

The two-RF Hahn-echo-like sequences closely matched the optimal grid-searched solution (88{\textdegree}--177{\textdegree}; Fig. \ref{fig4}a) in terms of flip angles, signal intensities, and RF energy. The slightly reduced flip angles resulted in lower RF energy (96.4 {\textpm} 0.2\% relative to the conventional 90{\textdegree}--180{\textdegree} sequence), which was nearly identical to that of the optimal solution (96.3\%), making these sequences optimal under the defined loss function.

The three-RF Hahn-echo-like sequences produced tissue signals comparable to the optimal design (GM/WM/CSF: 0.71 {\textpm} 0.00/0.60 {\textpm} 0.00/0.99 {\textpm} 0.00 for the three-RF design vs. 0.71/0.60/0.99 for the optimal design) and demonstrated similar robustness to $B_0$ and $B_1^+$ inhomogeneities (error = 2.2\% for the three-RF Hahn-echo-like sequences vs. 2.3\% for the optimal design; Fig. \ref{fig4}g, i). Notably, they required lower RF energy (90.4 {\textpm} 0.2\% relative to the conventional 90{\textdegree}--180{\textdegree} sequence) than the optimal solution (96.3\%). However, the three-RF Hahn-echo-like sequences resulted in a longer time from first RF to readout (TE = 18.9 {\textpm} 0.3 ms), compared with the two-RF Hahn-echo-like sequences and optimal design (TE = 12.4 {\textpm} 0.0 ms).

The remaining 34\% of the generated sequences were suboptimal in terms of the design objectives, displaying lower WM, GM and CSF intensities compared to the successful designs, often due to the absence of an effective refocusing pulse or the presence of unnecessary preparatory inversion pulses.

\subsection{Experiment 2: Enhance GM--WM contrast}
\subsubsection{Identical T\textsubscript{1} for gray matter and white matter}
\begin{figure}[!b]
\centerline{\includegraphics[width=\linewidth]{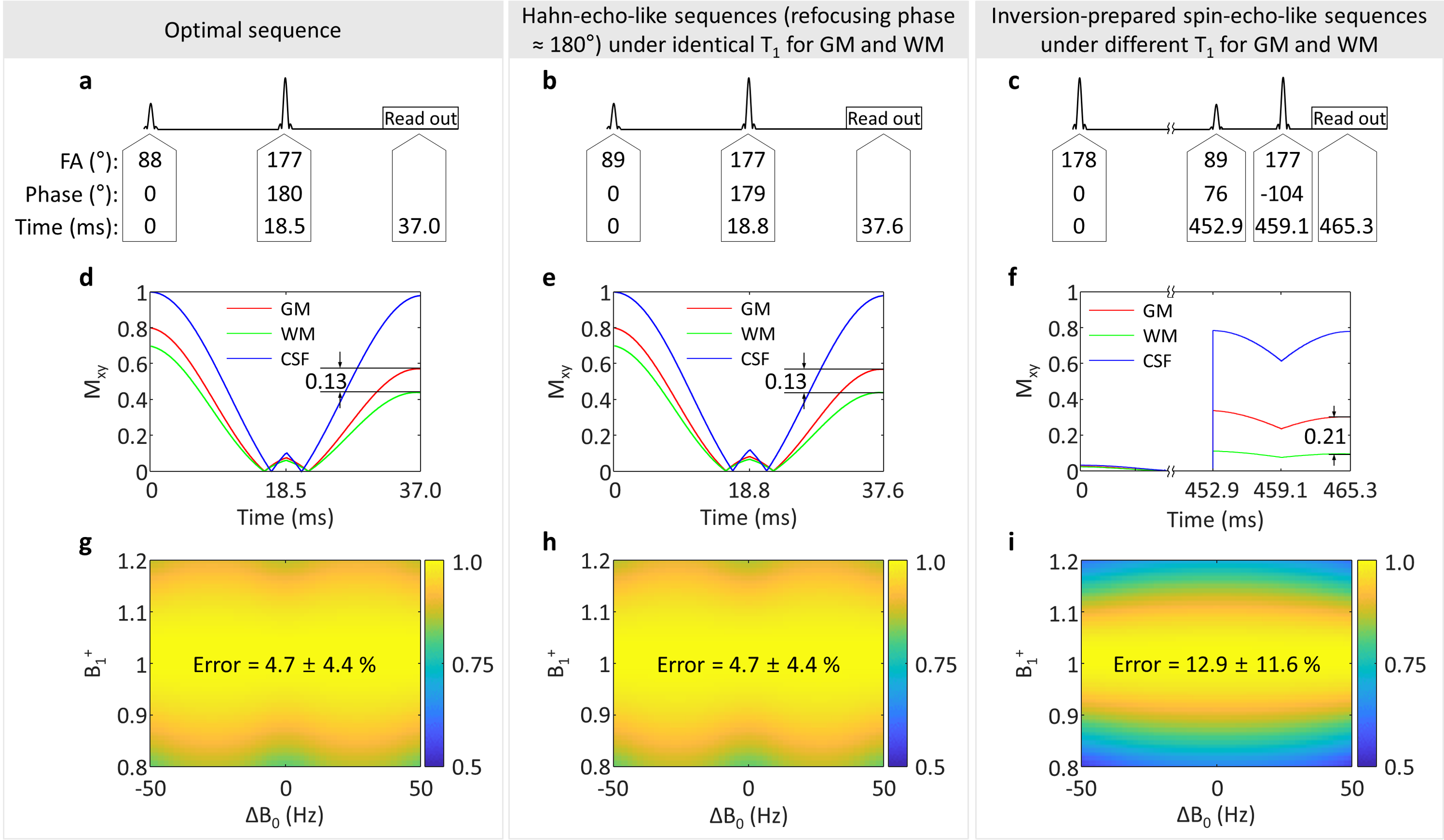}}
\caption{Results of the GM--WM contrast enhancement experiment (contrast weight = 30). (a) Optimal two-RF spin-echo identified by grid search (88{\textdegree}--177{\textdegree}, TE = 37.0 ms). (b, c) Designed sequences: (b) two-RF Hahn-echo-like sequence with a refocusing phase near 180{\textdegree} discovered under the identical-$T_1$ condition; (c) inversion-prepared spin-echo-like sequence discovered under the different-$T_1$ condition (TI = 452.9 ms). (d--f) Simulated transverse-magnetization evolution for GM, WM, and CSF corresponding to (a--c) at $B_1^+$=1 and $\Delta B_0$=0. (g--i) Robustness to field inhomogeneity: GM signal intensity over $\Delta B_0$ and $B_1^+$, normalized to the on-resonance value.
}
\label{fig5}
\end{figure}

Among the categorized sequences, Sequence Search dominantly generated Hahn-echo-like sequences with a refocusing phase near 180{\textdegree} ([88.6 {\textpm} 0.6{\textdegree}]--[177.3 {\textpm} 0.8{\textdegree}]; 40\% occurrence rate; Fig. \ref{fig5}b). These sequences had a TE of 37.6 {\textpm} 0.2 ms, matching the optimal grid-searched solution (TE = 37.0 ms; Fig. \ref{fig5}a). Compared to the optimal solution, the Hahn-echo-like sequences exhibited nearly identical signal intensities (GM/WM/CSF: 0.57 {\textpm} 0.00/0.44 {\textpm} 0.00/0.98 {\textpm} 0.00 vs. 0.57/0.44/0.98 for the optimal design) and contrast between GM and WM (0.13 {\textpm} 0.00 vs. 0.13 for the optimal design) with comparable RF energy (97.0 {\textpm} 0.9\% vs. 96.3\% for the optimal design).

\begin{table*}[!t]
\centering
\footnotesize
\caption{Results of the GM-WM contrast enhancement experiment (contrast weight = 30)}
\label{table6}
\setlength{\tabcolsep}{0pt}

\begin{tabular}{@{}
>{\centering\arraybackslash}m{0.10\textwidth}   
>{\centering\arraybackslash}m{0.20\textwidth}   
>{\centering\arraybackslash}m{0.08\textwidth}   
>{\centering\arraybackslash}m{0.05\textwidth}   
>{\centering\arraybackslash}m{0.05\textwidth}   
>{\centering\arraybackslash}m{0.07\textwidth}   
>{\centering\arraybackslash}m{0.05\textwidth}   
>{\centering\arraybackslash}m{0.09\textwidth}   
>{\centering\arraybackslash}m{0.12\textwidth}   
>{\centering\arraybackslash}m{0.10\textwidth}   
>{\centering\arraybackslash}m{0.09\textwidth}   
@{}}
\toprule

\multicolumn{2}{c}{\multirow{2}{*}{Category}} &
\multirow{2}{*}{\parbox[c]{0.08\textwidth}{\centering Occurrence rate (\%)}} &
\multicolumn{4}{c}{Signal intensity} &
\multirow{2}{*}{\parbox[c]{0.09\textwidth}{\centering Relative RF energy (\%)}} &
\multirow{2}{*}{\parbox[c]{0.12\textwidth}{\centering Time from the first RF to the readout (ms)}} &
\multirow{2}{*}{\parbox[c]{0.10\textwidth}{\centering Flip angles ({\textdegree})}} &
\multirow{2}{*}{\parbox[c]{0.09\textwidth}{\centering Refocusing RF phase ({\textdegree})}} \\

\cmidrule(lr){4-7}
& & & GM & WM & GM - WM & CSF & & & & \\

\midrule
\multicolumn{2}{c}{Conventional Hahn-echo sequence (90{\textdegree}--180{\textdegree})}
& -- & 0.49 & 0.36 & 0.13 & 0.97
& 100.0 & 54.2 & [90.0, 180.0] & 180.0 \\

\midrule

\multicolumn{2}{c}{Optimal grid-search sequence (88{\textdegree}--177{\textdegree})}
& -- & 0.57 & 0.44 & 0.13 & 0.98
& 96.3 & 37.0 & [88.0, 177.0] & 180.0 \\

\midrule

\multicolumn{11}{l}{\textbf{a) Identical T$_1$ for GM and WM}} \\

\midrule

\multirow{3}{*}{\parbox[c]{0.10\textwidth}{\centering Successful designs}}
& Hahn-echo-like sequences (refocusing phase $\approx$ 180{\textdegree})
& 40
& 0.57 {\textpm} 0.00 & 0.44 {\textpm} 0.00 & 0.13 {\textpm} 0.00 & 0.98 {\textpm} 0.00
& 97.0 {\textpm} 0.9 & 37.6 {\textpm} 0.2
& [88.6 {\textpm} 0.6, 177.3 {\textpm} 0.8]
& 179.3 {\textpm} 10.0 \\

& Hahn-echo-like sequences (refocusing phase $\approx$ 0{\textdegree})
& 3
& 0.54 {\textpm} 0.01 & 0.41 {\textpm} 0.01 & 0.13 {\textpm} 0.00 & 0.96 {\textpm} 0.01
& 97.7 {\textpm} 1.3 & 40.2 {\textpm} 0.2
& [88.9 {\textpm} 0.7, 177.9 {\textpm} 1.1]
& 4.1 {\textpm} 6.5 \\

& Other spin-echo-like sequences
& 16
& 0.56 {\textpm} 0.01 & 0.43 {\textpm} 0.01 & 0.13 {\textpm} 0.00 & 0.98 {\textpm} 0.01
& 97.3 {\textpm} 0.8 & 38.6 {\textpm} 1.3
& [88.9 {\textpm} 0.4, 177.6 {\textpm} 0.7]
& 111.0 {\textpm} 57.5 \\

\midrule

\multicolumn{2}{c}{Others}
& 41
& 0.50 {\textpm} 0.20 & 0.40 {\textpm} 0.16 & 0.11 {\textpm} 0.04 & 0.82 {\textpm} 0.30
& 156.6 {\textpm} 50.3 & 36.7 {\textpm} 6.0
& -- & -- \\

\midrule

\multicolumn{11}{l}{\textbf{b) Different T$_1$ for GM and WM}} \\

\midrule

\multirow{4}{*}{\parbox[c]{0.10\textwidth}{\centering Successful designs}}
& Hahn-echo-like sequences (refocusing phase $\approx$ 180{\textdegree})
& 42
& 0.57 {\textpm} 0.00 & 0.44 {\textpm} 0.00 & 0.13 {\textpm} 0.00 & 0.98 {\textpm} 0.00
& 97.2 {\textpm} 1.0 & 37.6 {\textpm} 0.2
& [88.7 {\textpm} 0.7, 177.4 {\textpm} 0.8]
& 178.1 {\textpm} 9.5 \\

& Hahn-echo-like sequences (refocusing phase $\approx$ 0{\textdegree})
& 2
& 0.55 {\textpm} 0.00 & 0.42 {\textpm} 0.00 & 0.13 {\textpm} 0.00 & 0.98 {\textpm} 0.00
& 97.7 {\textpm} 0.4 & 40.5 {\textpm} 0.1
& [88.8 {\textpm} 0.6, 178.0 {\textpm} 0.1]
& 1.7 {\textpm} 23.1 \\

& Other spin-echo-like sequences
& 12
& 0.56 {\textpm} 0.01 & 0.43 {\textpm} 0.01 & 0.13 {\textpm} 0.00 & 0.98 {\textpm} 0.01
& 97.2 {\textpm} 1.0 & 38.4 {\textpm} 1.0
& [88.7 {\textpm} 0.6, 177.4 {\textpm} 0.9]
& 121.7 {\textpm} 56.6 \\

& Inversion-prepared spin-echo-like sequences
& 17
& 0.30 {\textpm} 0.00 & 0.10 {\textpm} 0.00 & 0.21 {\textpm} 0.00 & 0.78 {\textpm} 0.00
& 174.9 {\textpm} 1.4 & 465.3 {\textpm} 5.4
& [178.1 {\textpm} 1.0, 88.5 {\textpm} 0.3, 176.9 {\textpm} 0.9]
& 179.9 {\textpm} 0.5 \\

\midrule

\multicolumn{2}{c}{Others}
& 27
& 0.25 {\textpm} 0.24 & 0.23 {\textpm} 0.17 & 0.13 {\textpm} 0.08 & 0.60 {\textpm} 0.31
& 139.2 {\textpm} 61.6 & 393.4 {\textpm} 423.7
& -- & -- \\

\bottomrule
\end{tabular}
\end{table*}
\begin{table*}[!t]
\centering
\footnotesize
\caption{Parameters of the most frequently observed (highest occurrence) two-RF spin-echo-like sequences for each contrast weighting coefficient}
\label{table7}
\setlength{\tabcolsep}{0pt}

\begin{tabular}{@{}
>{\centering\arraybackslash}m{0.10\textwidth}
>{\centering\arraybackslash}m{0.10\textwidth}
>{\centering\arraybackslash}m{0.10\textwidth}
>{\centering\arraybackslash}m{0.10\textwidth}
>{\centering\arraybackslash}m{0.10\textwidth}
>{\centering\arraybackslash}m{0.18\textwidth}
>{\centering\arraybackslash}m{0.18\textwidth}
>{\centering\arraybackslash}m{0.14\textwidth}
@{}}
\toprule

\multirow{2}{*}{\parbox[c]{0.10\textwidth}{\centering Contrast weight}} &
\multirow{2}{*}{\parbox[c]{0.10\textwidth}{\centering Occurrence rate (\%)}} &
\multicolumn{3}{c}{Signal intensity} &
\multirow{2}{*}{\parbox[c]{0.18\textwidth}{\centering Time from the first RF to the readout (ms)}} &
\multirow{2}{*}{\parbox[c]{0.18\textwidth}{\centering Flip angles ({\textdegree})}} &
\multirow{2}{*}{\parbox[c]{0.14\textwidth}{\centering Refocusing RF phase ({\textdegree})}} \\

\cmidrule(lr){3-5}
& & GM & WM & GM - WM & & & \\

\midrule

20 & 25
& 0.66 {\textpm} 0.06
& 0.54 {\textpm} 0.07
& 0.12 {\textpm} 0.01
& 21.9 {\textpm} 10.3
& [88.8 {\textpm} 0.7, 177.5 {\textpm} 1.0]
& 178.9 {\textpm} 4.9 \\

\midrule

30 & 40
& 0.57 {\textpm} 0.00
& 0.44 {\textpm} 0.00
& 0.13 {\textpm} 0.00
& 37.6 {\textpm} 0.2
& [88.6 {\textpm} 0.6, 177.3 {\textpm} 0.8]
& 179.3 {\textpm} 10.0 \\

\midrule

50 & 43
& 0.53 {\textpm} 0.00
& 0.40 {\textpm} 0.00
& 0.13 {\textpm} 0.00
& 45.0 {\textpm} 0.2
& [88.9 {\textpm} 0.8, 177.6 {\textpm} 0.9]
& 1.3 {\textpm} 8.5 \\

\bottomrule
\end{tabular}
\end{table*}

A smaller subset of Hahn-echo-like sequences with a refocusing phase near 0{\textdegree} was also observed (3\% occurrence rate). These sequences exhibited slightly lower signal intensities (GM/WM/CSF: 0.54 {\textpm} 0.01/0.41 {\textpm} 0.01/0.96 {\textpm} 0.01), similar contrast (0.13 {\textpm} 0.00), RF energy (97.7 {\textpm} 1.3\%), and TE (40.2 {\textpm} 0.2 ms) compared to the optimal design. 16\% of the sequences were classified as other spin-echo-like configurations ([88.9 {\textpm} 0.4{\textdegree}]--[177.6 {\textpm} 0.7{\textdegree}]). These designs achieved signals (GM/WM/CSF: 0.56 {\textpm} 0.01/0.43 {\textpm} 0.01/0.98 {\textpm} 0.01), contrast (0.13 {\textpm} 0.00), RF energy (97.3 {\textpm} 0.8\%), and TE (38.6 {\textpm} 1.3 ms) comparable to the optimal design.

The remaining sequences (41\%) showed degraded signal contrast, higher RF energy, or suboptimal timing and were therefore categorized as Others.

\subsubsection{Different T\textsubscript{1} for gray matter and white matter}
When GM and WM had different $T_1$ values, Sequence Search generated several sequence types (Table \ref{table6}-b), including Hahn-echo-like sequences, other spin-echo-like sequences, and inversion-prepared spin-echo-like sequences. The dominant solutions were Hahn-echo-like sequences with refocusing phases near 180{\textdegree} ([88.7 {\textpm} 0.7{\textdegree}]--[177.4 {\textpm} 0.8{\textdegree}]; 42\% occurrence rate; TE = 37.6 {\textpm} 0.2 ms). These sequences exhibited signal intensities (GM/WM/CSF: 0.57 {\textpm} 0.00/0.44 {\textpm} 0.00/0.98 {\textpm} 0.00), contrast (0.13 {\textpm} 0.00), RF energy (97.2 {\textpm} 1.0\%), and TE (37.6 {\textpm} 0.2 ms) comparable to those of the optimal solution.

A small subset of Hahn-echo-like sequences with refocusing phases near 0{\textdegree} ([88.8 {\textpm} 0.6{\textdegree}]--[178.0 {\textpm} 0.1{\textdegree}]; 2\% occurrence rate) was also observed, exhibiting similar signal intensities (GM/WM/CSF: 0.55 {\textpm} 0.00/0.42 {\textpm} 0.00/0.98 {\textpm} 0.00), contrast (0.13 {\textpm} 0.00), RF energy (97.7 {\textpm} 0.4\%), and TE (40.5 {\textpm} 0.1 ms). 12\% of the sequences were classified as other spin-echo-like sequences ([88.7 {\textpm} 0.6{\textdegree}]--[177.4 {\textpm} 0.9{\textdegree}]). These sequences produced signal intensities (GM/WM/CSF: 0.56 {\textpm} 0.01/0.43 {\textpm} 0.01/0.98 {\textpm} 0.01), GM--WM contrast (0.13 {\textpm} 0.00), RF energy (97.2 {\textpm} 1.0\%), and TE (38.4 {\textpm} 1.0 ms) comparable to the optimal solution.

A distinct class of inversion-prepared spin-echo-like sequences (17\% occurrence rate) was also identified ([178.1 {\textpm} 1.0{\textdegree}]--[88.5 {\textpm} 0.3{\textdegree}]--[176.9 {\textpm} 0.9{\textdegree}]; Fig. \ref{fig5}c). Inversion-prepared spin-echo-like sequences, with a TI of 459.1 {\textpm} 5.4 ms, produced a larger GM--WM signal difference (0.21 {\textpm} 0.00), exceeding that of the Hahn-echo-like sequences (0.13 {\textpm} 0.00). However, this improvement came at the cost of substantially longer time from the first RF to readout (465.3 {\textpm} 5.4 ms vs. 37.6 {\textpm} 0.2 ms) and reduced signal intensities (GM/WM/CSF: 0.30 {\textpm} 0.00/0.10 {\textpm} 0.00/0.78 {\textpm} 0.00 vs. 0.57 {\textpm} 0.00/0.44 {\textpm} 0.00/0.98 {\textpm} 0.00) compared to the Hahn-echo-like sequences. These sequences also showed greater sensitivity to field inhomogeneities (12.9\% error; Fig. \ref{fig5}i).

\begin{table*}[!t]
\centering
\footnotesize
\caption{Results of the CSF nulling experiment}
\label{table8}
\setlength{\tabcolsep}{0pt}

\begin{tabular}{@{}
>{\centering\arraybackslash}m{0.11\textwidth}   
>{\centering\arraybackslash}m{0.24\textwidth}   
>{\centering\arraybackslash}m{0.09\textwidth}   
>{\centering\arraybackslash}m{0.09\textwidth}   
>{\centering\arraybackslash}m{0.09\textwidth}   
>{\centering\arraybackslash}m{0.09\textwidth}   
>{\centering\arraybackslash}m{0.09\textwidth}   
>{\centering\arraybackslash}m{0.10\textwidth}   
>{\centering\arraybackslash}m{0.10\textwidth}   
@{}}
\toprule

\multicolumn{2}{c}{\multirow{2}{*}{Category}} &
\multirow{2}{*}{\parbox[c]{0.09\textwidth}{\centering Occurrence rate (\%)}} &
\multicolumn{3}{c}{Signal intensity} &
\multirow{2}{*}{\parbox[c]{0.09\textwidth}{\centering Relative RF energy (\%)}} &
\multirow{2}{*}{\parbox[c]{0.10\textwidth}{\centering Time from the first RF to the readout (s)}} &
\multirow{2}{*}{\parbox[c]{0.10\textwidth}{\centering Flip angles ({\textdegree})}} \\

\cmidrule(lr){4-6}
& & & GM & WM & CSF & & & \\

\midrule

\multicolumn{2}{c}{Conventional inversion recovery sequence (180{\textdegree}--90{\textdegree})}
& -- & 0.40 & 0.42 & 0.00
& 100.0 & 2.78 & [180.0, 90.0] \\

\midrule

\multicolumn{9}{l}{\textbf{a) CSF nulling weight = 0.25}} \\

\midrule

\multicolumn{2}{c}{Optimal grid-search sequence (176{\textdegree}--89{\textdegree})}
& -- & 0.49 & 0.47 & 0.17
& 96.0 & 3.75 & [176.0, 89.0] \\

\midrule

\multirow{3}{*}{\parbox[c]{0.11\textwidth}{\centering Successful designs}}
& Inversion-prepared gradient-echo-like sequences (176{\textdegree}--89{\textdegree})
& 47
& 0.49 {\textpm} 0.00 & 0.47 {\textpm} 0.00 & 0.17 {\textpm} 0.01
& 96.1 {\textpm} 0.4 & 3.77 {\textpm} 0.04
& [176.1 {\textpm} 0.2, 88.7 {\textpm} 0.1] \\

& Inversion-prepared spin-echo-like sequences (177{\textdegree}--88{\textdegree}--176{\textdegree})
& 19
& 0.60 {\textpm} 0.00 & 0.58 {\textpm} 0.00 & 0.13 {\textpm} 0.00
& 172.9 {\textpm} 0.2 & 3.35 {\textpm} 0.01
& [176.6 {\textpm} 0.2, 88.1 {\textpm} 0.1, 176.2 {\textpm} 0.1] \\

\midrule

\multicolumn{2}{c}{Others}
& 34
& 0.42 {\textpm} 0.19 & 0.38 {\textpm} 0.18 & 0.24 {\textpm} 0.22
& 154.9 {\textpm} 54.7 & 2.82 {\textpm} 2.28
& -- \\

\midrule

\multicolumn{9}{l}{\textbf{b) CSF nulling weight = 1.0}} \\

\midrule

\multicolumn{2}{c}{Optimal grid-search sequence (178{\textdegree}--89{\textdegree})}
& -- & 0.45 & 0.45 & 0.06
& 97.8 & 3.07 & [178.0, 89.0] \\

\midrule

\multirow{3}{*}{\parbox[c]{0.11\textwidth}{\centering Successful designs}}
& Inversion-prepared gradient-echo-like sequences (178{\textdegree}--89{\textdegree})
& 19
& 0.45 {\textpm} 0.00 & 0.45 {\textpm} 0.00 & 0.06 {\textpm} 0.00
& 97.2 {\textpm} 0.1 & 3.07 {\textpm} 0.01
& [177.5 {\textpm} 0.2, 88.7 {\textpm} 0.0] \\

& Inversion-prepared spin-echo-like sequences (178{\textdegree}--88{\textdegree}--176{\textdegree})
& 9
& 0.55 {\textpm} 0.00 & 0.56 {\textpm} 0.00 & 0.03 {\textpm} 0.00
& 174.0 {\textpm} 0.1 & 2.91 {\textpm} 0.01
& [177.8 {\textpm} 0.1, 88.2 {\textpm} 0.1, 176.3 {\textpm} 0.1] \\

\midrule

\multicolumn{2}{c}{Others}
& 72
& 0.05 {\textpm} 0.13 & 0.05 {\textpm} 0.13 & 0.02 {\textpm} 0.05
& 147.9 {\textpm} 65.0 & 4.89 {\textpm} 2.22
& -- \\

\bottomrule
\end{tabular}
\end{table*}

\subsubsection{Effect of contrast weighting}
Across contrast weights ($\lambda_\textrm{cont}$ = 20, 30, 50) under the identical $T_1$ setting, Sequence Search consistently produced two-RF spin-echo solutions with flip angles close to the conventional 90{\textdegree}--180{\textdegree} configuration (Table \ref{table7}). The parameter values reported in Table \ref{table7} correspond to the most frequently observed (highest occurrence) solutions for each contrast weighting.

For $\lambda_\textrm{cont}$ = 20, the designed sequences had TE of 21.9 {\textpm} 10.3 ms, producing GM and WM signals of 0.66 {\textpm} 0.06 and 0.54 {\textpm} 0.07, respectively, corresponding to a GM--WM difference of 0.12 {\textpm} 0.01. Increasing the contrast weight to 30 resulted in TE of 37.6 {\textpm} 0.2 ms with GM and WM signals of 0.57 {\textpm} 0.00 and 0.44 {\textpm} 0.00, yielding a GM--WM difference of 0.13 {\textpm} 0.00. For $\lambda_\textrm{cont}$ = 50, the designed sequences had TE of 45.0 {\textpm} 0.2 ms with GM and WM signals of 0.53 {\textpm} 0.00 and 0.40 {\textpm} 0.00, producing a GM--WM difference of 0.13 {\textpm} 0.00.

The preferred refocusing RF phase of the dominant solutions depended on the contrast weighting: phases near 180{\textdegree} were observed for $\lambda_\textrm{cont}$ = 20 and 30, whereas for  $\lambda_\textrm{cont}$ = 50, the dominant solutions exhibited a phase near 0{\textdegree}.

\subsection{Experiment 3: CSF nulling}
\begin{figure}[!b]
\centerline{\includegraphics[width=\linewidth]{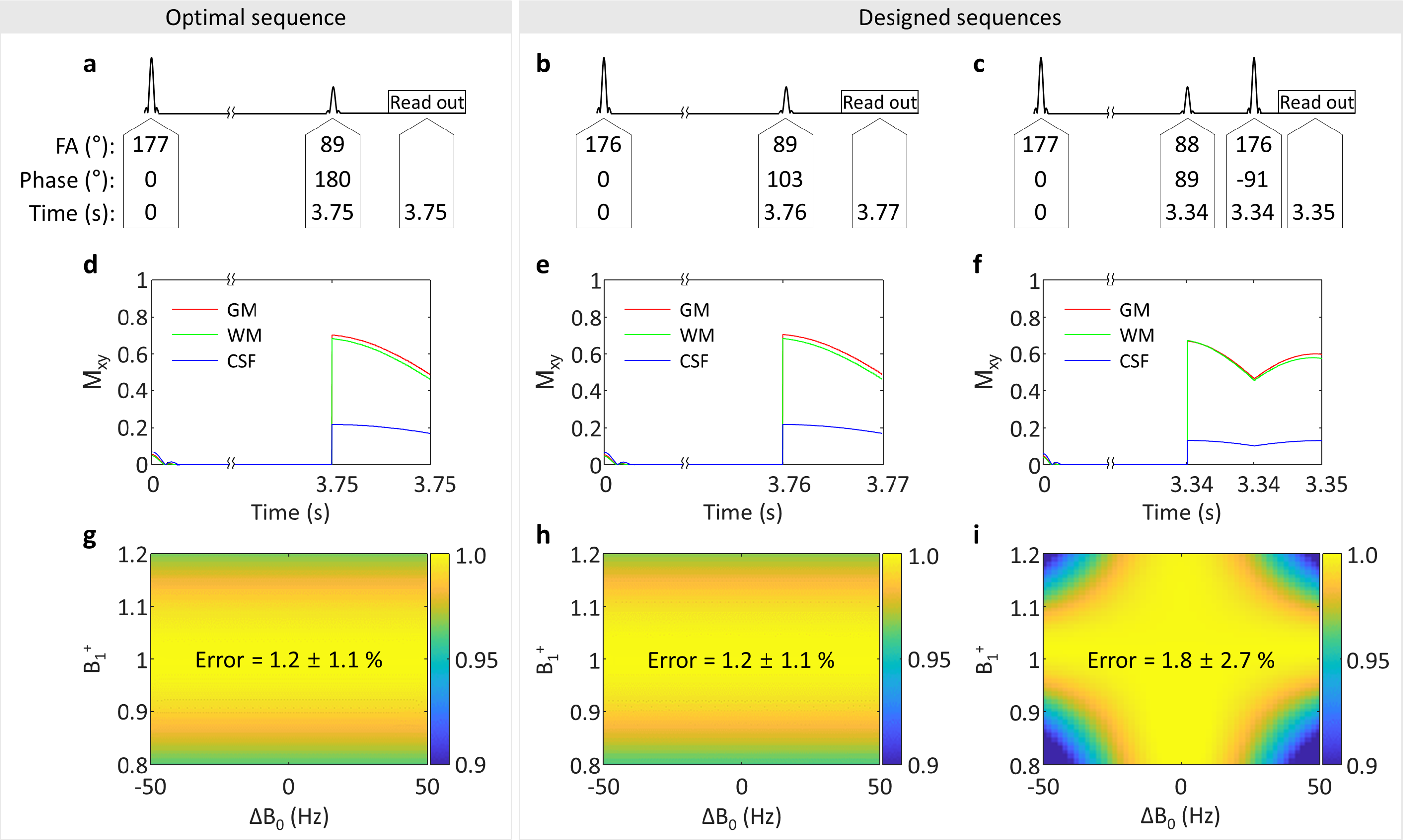}}
\caption{Results of the CSF nulling experiment (CSF nulling weight = 0.25). (a) Optimal inversion recovery sequence identified by grid search (177{\textdegree}–89{\textdegree}; TI = 3.75 s). (b, c) Designed sequences: (b) inversion-prepared gradient-echo-like sequence (TI = 3.76 s); (c) inversion-prepared spin-echo-like sequence (TI = 3.34 s). (d--f) Simulated transverse-magnetization evolution for GM, WM, and CSF corresponding to (a--c) at $B_1^+$=1 and $\Delta B_0$=0. (g--i) Robustness to field inhomogeneity: GM signal intensity over $\Delta B_0$ and $B_1^+$, normalized to the on-resonance value.
}
\label{fig6}
\end{figure}

With CSF-nulling weight of 0.25, Sequence Search generated two dominant sequence types capable of suppressing CSF signal while preserving GM and WM signals (Table \ref{table8}-a): inversion-prepared gradient-echo-like sequences ([176.1 {\textpm} 0.2{\textdegree}]--[88.7 {\textpm} 0.1{\textdegree}]; 47\% occurrence rate; Fig. \ref{fig6}b) and inversion-prepared spin-echo-like sequences ([176.6 {\textpm} 0.2{\textdegree}]--[88.1 {\textpm} 0.1{\textdegree}]--[176.2 {\textpm} 0.1{\textdegree}]; 19\% occurrence rate; Fig. \ref{fig6}c).

Inversion-prepared gradient-echo-like sequences exhibited TI of 3.77 {\textpm} 0.04 s, closely matching the optimal grid-searched solution (TI = 3.75 s; Fig. \ref{fig6}a). These designs yielded GM and WM signals comparable to the optimal design (GM/WM = 0.49 {\textpm} 0.00/0.47 {\textpm} 0.00 for these designs vs. 0.49/0.47 for the optimal design), with residual CSF signal also comparable to the optimal design (0.17 {\textpm} 0.01 for these designs vs. 0.17 for the optimal design), although remaining higher than that of the conventional inversion recovery sequence (0.00). Notably, they required lower RF energy (96.1 {\textpm} 0.4\%) than the conventional inversion recovery sequence, and demonstrated 1.2\% signal variation under simulated field inhomogeneities (Fig. \ref{fig6}h).

Inversion-prepared spin-echo-like sequences, with TI = 3.34 {\textpm} 0.01 s, produced higher GM and WM signals (GM/WM: 0.60 {\textpm} 0.00/0.58 {\textpm} 0.00 for these designs vs. 0.49/0.47 for the optimal design) and stronger CSF suppression (0.13 {\textpm} 0.00 for these designs vs. 0.17 for the optimal design). However, these gains were accompanied by substantially higher RF energy consumption (172.9 {\textpm} 0.2\% relative to the conventional sequence) and reduced robustness to field inhomogeneities (1.8\% error for these designs vs. 1.2\% error for the optimal design; Fig. \ref{fig6}i vs. Fig. \ref{fig6}g). The remaining 34\% of generated sequences were categorized as Others.

When the CSF-nulling weight was increased to 1.0 (Table \ref{table8}-b), designed sequences shifted toward more aggressive CSF suppression. Inversion-prepared gradient-echo-like sequences (19\% occurrence rate) exhibited a shorter TI (3.07 {\textpm} 0.01 s), approaching the theoretical CSF nulling point (2.77 s), and achieved lower residual CSF signal (0.06 {\textpm} 0.00). RF energy remained comparable to the conventional design (97.2 {\textpm} 0.1\%). Inversion-prepared spin-echo-like sequences also emerged (9\% occurrence rate), with a TI of 2.91 {\textpm} 0.01 s. These designs produced higher GM and WM signals (GM/WM: 0.55 {\textpm} 0.00/0.56 {\textpm} 0.00) and further reduced CSF signal (0.03 {\textpm} 0.00) compared to the gradient-echo-based designs, but at the cost of substantially increased RF energy (174.0 {\textpm} 0.1\%).

\section{Discussion}
\subsection{Refocusing phase and quasi-optimal solutions}
Across Experiments 1 and 2, Sequence Search predominantly generated spin-echo sequences with refocusing phases near 0{\textdegree} or 180{\textdegree}, corresponding to conventional Hahn-echo configurations. This behavior follows from the optimization objective, which targeted a single echo robust to $B_0$ and $B_1^+$ variations; under these conditions, refocusing phases of 0{\textdegree} or 180{\textdegree} provide more robust echo formation than CPMG-type {\textpm}90{\textdegree} phases across the simulated $\Delta B_0$ and $B_1^+$ ranges.

In Experiment 2, the preferred refocusing phase depended on the contrast weighting used in the objective function. For contrast weights of 20 and 30, the designed sequences had refocusing phases near 180{\textdegree}. When the contrast weight increased to 50, however, the refocusing phase shifted toward 0{\textdegree}. These values match the optimal sequences identified independently under the corresponding objective settings (data not shown), indicating that the optimizer correctly captured the phase conditions that minimize the loss for each weighting scheme.

Sequence Search also produced solutions with diverse RF phases, likely reflecting a flat loss landscape where multiple phase configurations yield comparable signal characteristics.

\subsection{Parameter shifts from multi-objective optimization}
Although the discovered sequence structures corresponded to known spin-echo or inversion recovery configurations, the designed timing parameters (TE and TI) differed from conventional values derived from single-criterion signal models (e.g., TI = 2.78 s for CSF nulling). These shifts arise because Sequence Search simultaneously balances multiple objectives, including signal intensity, tissue contrast, RF energy, and robustness to field inhomogeneity.

For example, increasing the contrast weight in Experiment 2 resulted in longer TE (from $\sim$22 ms to $\sim$45 ms; Table \ref{table7}), reflecting the trade-off between signal intensity and GM–WM contrast. Similarly, in Experiment 3, the designed TI ($\sim$3.75 s at CSF nulling weight = 0.25) exceeded the theoretical CSF zero-crossing ($\sim$2.77 s) but shifted toward it ($\sim$3.07 s) when the CSF nulling weight was increased to 1.0. These observations indicate that the optimized parameters are determined by the relative weighting of competing objectives rather than by a single analytic optimum.

\subsection{Designs based on diverse contrast mechanisms}
In Experiment 1, Sequence Search identified three-RF spin-echo-like sequences that deviated from conventional two-RF Hahn-echo configurations. These sequences incorporated an additional low flip-angle RF pulse prior to the excitation and refocusing, resulting in similar signal behavior with substantially reduced RF energy. While the fundamental spin-echo refocusing mechanism is unaltered, the designed sequences display a new contrast mechanism that is difficult for human intuition to easily anticipate.

When GM and WM were assigned different $T_1$ values in Experiment 2, Sequence Search generated two types of sequences: Hahn-echo-like and inversion-prepared spin-echo sequences. While the spin-echo sequences rely on $T_2$ differences, the inversion-prepared sequences also exploit $T_1$ recovery differences. In contrast, inversion-prepared sequences did not emerge when GM and WM were assigned the same $T_1$ values. These results reflect the optimizer’s ability to exploit available tissue-contrast mechanisms depending on the specified tissue parameters.

While the inversion-prepared sequences demonstrated larger GM--WM contrast compared to the spin-echo designs, they also required extended time from first RF to readout, and displayed reduced signal magnitude. This demonstrates that the loss landscape, determined by a combination of multiple loss functions, may include multiple local optima, resulting in diverse sequence designs.

\subsection{Limitations and extensions of the current framework}
The current implementation simplified the search space to illustrate the core concept. Sequence design was limited to RF pulse scheduling, while gradient schemes were not explicitly modeled. In principle, gradient spoiling or slice-profile effects could be incorporated by introducing additional selectable operations and spatial dimensions. Similarly, the assumption of instantaneous RF pulses could be relaxed by incorporating RF waveforms as trainable parameters, enabling joint optimization of waveform shape and sequence timing or selection among predefined pulse types (e.g., sinc-shaped or adiabatic pulses).

The use of infinite TR isolated the intrinsic contrast behavior of a single sequence execution but did not capture steady-state effects under finite TR. Incorporating realistic TR constraints would require repeated Bloch simulations to compute steady-state magnetization, substantially increasing computational cost, although strategies such as approximate steady-state modeling or surrogate models could mitigate this.

Finally, the discovered sequences depend strongly on the loss function formulation. Alternative objective functions---such as explicit penalties on TE or scan duration, or metrics targeting novel contrasts---could lead to different and potentially clinically relevant sequence designs.

\subsection{Potential applications and relation to prior work}
Recent work such as MR-double-zero \cite{bib30} has explored autonomous discovery of MRI contrasts through iterative interaction with a scanner when the contrast mechanism is unknown. MR-double-zero optimizes parameters within a fixed sequence structure using model-free exploration. In contrast, the proposed method focuses on discovering the sequence structure itself and optimizing its parameters within a Bloch-simulation framework.

This capability may be particularly useful for applications such as microstructure-specific or biomarker-sensitive contrasts, where signal behavior can be difficult to derive analytically. By expanding the scope of the framework to more comprehensive sequence models and simulation environments, it may become possible to identify non-intuitive sequence designs that exploit specific physical tissue properties.

\section{Conclusion}
In conclusion, we introduced a novel automated sequence design method utilizing NAS, termed ``Sequence Search.’’ The NAS-based sequence scheduler generates a sequence, and this sequence is validated against a predefined loss function derived from the design objectives. Remarkably, this method produced sequences tailored to design objectives without relying on prior knowledge or an initial sequence structure. Our approach successfully replicated the design of conventional spin-echo and inversion recovery sequences, mirroring human intuition by reproducing well-known sequences. Furthermore, our method produced less-intuitive designs, such as spin-echo-like sequences using three RF pulses but with reduced RF energy. This study illuminates the potential of automated sequence designs that may explore possibilities beyond human intuition.

\section*{References}
\providecommand{\refname}{References}

\end{document}